\documentclass{article}

\usepackage{PRIMEarxiv}

\usepackage[utf8]{inputenc} 
\usepackage[T1]{fontenc}    
\usepackage{hyperref}       
\usepackage{url}            
\usepackage{booktabs}       
\usepackage{amsfonts}       
\usepackage{nicefrac}       
\usepackage{microtype}      
\usepackage{lipsum}
\usepackage{fancyhdr}       
\usepackage{graphicx}       
\graphicspath{{media/}}     

\usepackage{amsmath}
\usepackage{subcaption}
\usepackage[ruled,vlined]{algorithm2e}
\usepackage{authblk}

\title{Towards Efficient Multi-Objective Optimisation for Real-World Power Grid Topology Control
}

\author[1]{Yassine El Manyari}
\author[1]{Anton R. Fuxjäger}
\author[1]{Stefan Zahlner}
\author[2]{Joost Van Dijk}
\author[1]{Alberto Castagna}
\author[2]{Davide Barbieri}
\author[2]{Jan Viebahn}
\author[1]{Marcel Wasserer *}

\affil[1]{enliteAI, Vienna, Austria}
\affil[2]{TenneT TSO, Arnhem, The Netherlands}
\affil[*]{Corresponding author: m.wasserer@enlite.ai}

\begin{document}
\maketitle

\begin{abstract}
    Power grid operators face increasing difficulties in the control room as the increase in energy demand and the shift to renewable energy introduce new complexities in managing congestion and maintaining a stable supply. Effective grid topology control requires advanced tools capable of handling multi-objective trade-offs. While Reinforcement Learning (RL) offers a promising framework for tackling such challenges, existing Multi-Objective Reinforcement Learning (MORL) approaches fail to scale to the large state and action spaces inherent in real-world grid operations. Here we present a two-phase, efficient and scalable Multi-Objective Optimisation (MOO) method designed for grid topology control, combining an efficient RL learning phase with a rapid planning phase to generate day-ahead plans for unseen scenarios. We validate our approach using historical data from TenneT, a European Transmission System Operator (TSO), demonstrating minimal deployment time, generating day-ahead plans within 4–7 minutes with strong performance. These results underline the potential of our scalable method to support real-world power grid management, offering a practical, computationally efficient, and time-effective tool for operational planning. Based on current congestion costs and inefficiencies in grid operations, adopting our approach by TSOs could potentially save millions of euros annually, providing a compelling economic incentive for its integration in the control room.
\end{abstract}


\section{Introduction}
    The increasing complexity and evolving demand of modern electric grids present a critical challenge to operators worldwide \cite{pillayCongestionManagementPower2015}. Rapid electrification in industry, buildings, and transportation, combined with the integration of more renewable energy sources—often intermittent and geographically dispersed—pushes current infrastructures to their limits \cite{viebahnGridOptionsToolRealWorld2024}. These shifts lead to increased congestion, more complex maintenance, and higher operational costs. To navigate such complexities, Transmission System Operators (TSOs) need advanced decision-support tools capable of dynamically and efficiently optimising grid topologies. 
 
    Multi-Objective Optimisation (MOO) is central to this endeavour, as TSOs often face several competing priorities simultaneously. For example, they may strive to minimise transmission line loads, reduce the frequency of switching actions, or limit the topological complexity of the network \cite{viebahnGridOptionsToolRealWorld2024}. Each of these objectives, which we will detail later, serves as an example of the diverse performance criteria that must be balanced. Meeting these objectives using conventional methods can be computationally daunting, especially given the large action and state spaces associated with real-world grids \cite{viebahnGridOptionsToolRealWorld2024}. 
    
    Reinforcement Learning (RL) provides a promising framework for discovering effective strategies in such complex environments \cite{damjanovicDeepReinforcementLearningBased2022, bernadicReinforcementLearningPower2023, landenDRAGONDeepReinforcement2022, lehnaManagingPowerGrids2023a, dorferPowerGridCongestion2022}. However, accommodating multiple objectives simultaneously remains a significant challenge. Multi-Objective Reinforcement Learning (MORL) algorithms aim to address this by enabling agents to learn policies for different trade-offs between various objectives \cite{hayesPracticalGuideMultiobjective2022}. Yet, state-of-the-art MORL solutions fail to scale effectively for real-world grid optimisation problems due to the large state and action spaces, making them unsuitable for handling complex grids with numerous possible configurations.
    
    In this paper, we propose a novel MOO approach for grid topology optimisation that is both computationally efficient and time-effective. Our approach aims to overcome the limitations of current MORL methods, successfully managing expansive action and state spaces without sacrificing performance or scalability. By doing so, we bring MOO closer to real-world power grid topology control for operational planning.

\section{Background}
\label{section:background}

    \textbf{Electric Grid Representation.}  
        A power grid can be represented as an undirected graph $ G = (V, E) $, where $ V $ denotes the set of nodes (substations) and $ E $ represents the set of edges (transmission lines). Each substation $ v \in V $ is equipped with two buses ($ A, B $) to which transmission lines ($ e \in E $) can be connected. Referred to collectively as injections, generators and loads are directly connected to these buses. Part of the grid used in our experiments is depicted in Fig.~\ref{fig:grid_network} in the Appendix. At any given time $ t $, the total power generation must satisfy the total power demand across the grid, and each transmission line $ e \in E $ has a thermal limit $ P_{\text{max}}(e) $, beyond which the grid's stability may be compromised. TSOs continuously monitor the grid and execute remedial actions to maintain safe operational conditions. These actions include reconfiguring the grid topology at substations (e.g., switching lines between buses) and adjusting generator outputs.
    
    \textbf{Reinforcement Learning (RL).}  Power grid operations can be formulated as a sequential decision-making problem, where decisions are made at discrete time steps to optimise the grid's performance. At each time step $ t $, the state of the grid is represented by a set of variables collectively called the state $ s_t $. These variables can include generators output, line flows, and other grid parameters. The operator selects an action $ a_t $, such as altering substation configurations or redispatching generators, which transitions the grid to a new state $ s_{t+1} $. In this work, we only consider grid topology/configuration actions.
        
        RL provides a powerful framework to solve this sequential decision-making problem. In RL, an agent learns an optimal policy $ \pi(a_t | s_t) $, which maps states to actions to maximise cumulative rewards over a time horizon. The reward $r(s_t, a_t)$ is a scalar signal provided by the environment that quantifies the immediate benefit of taking a particular action in a given state. By exploring different actions and evaluating the resulting rewards, the agent incrementally refines its decision-making strategy. The optimisation objective for the RL agent can be expressed as: $\max_{\pi} \mathbb{E}\left[\sum_{t=0}^{T} \gamma^t r(s_t, a_t)\right]$, where $ \gamma \in [0, 1] $ is a discount factor that balances immediate and future rewards. More details about RL can be found in \cite{suttonReinforcementLearningIntroduction1998}.

    \textbf{Multi-Objective Optimisation (MOO).}  In many real-world grid operation problems, multiple conflicting objectives must be balanced simultaneously. For example, TenneT TSO in \cite{viebahnGridOptionsToolRealWorld2024}, considers the following objectives in power grid management:

    \begin{enumerate}
        \item Minimise the N-0 load flow.
        \item Minimise the N-1 load flow.
        \item Minimise the number of switching timestamps.
        \item Minimise the number of open busbar couplers (i.e. minimise the topological depth).
        \item Minimise the topological distance between consecutive topologies.
    \end{enumerate}

    These objectives capture different priorities, ranging from ensuring stability under normal and contingency conditions (N-0 and N-1) to reducing operational complexity and minimising frequent reconfiguration. In this work, we focus on objectives (2) and (3). Optimising objective (2) ensures resilience under single-contingency scenario (N-1) and implicitly optimises N-0 loads as well, resulting in decreased maintenance, longer durability of power lines and reduced energy losses. Optimising objective (3) decreases the frequency of grid topology changes required by operators within a single day.

    The Pareto Front (PF) is a key concept in multi-objective optimisation. It represents the set of solutions that are Pareto optimal, meaning that no other solution can improve one objective without degrading at least one other objective. In the context of grid operations, the PF provides the set of optimal trade-offs among objectives, allowing grid operators to select policies that align with their priorities. The ultimate aim of MOO is to approximate this PF.
    
    \textbf{MOO Metrics.}  
    Evaluating MOO solutions typically involves metrics that assess the quality of policies in multi-objective space. In this work, we use the hypervolume metric, which measures the volume in the objective space dominated by the solution set relative to a fixed reference point. A larger hypervolume indicates better spread and quality of solutions that approximate the PF \cite{roijersSurveyMultiObjectiveSequential2013}. For a comprehensive review of MOO concepts and metrics, see \cite{hayesPracticalGuideMultiobjective2022}.

\section{Problem Formulation and Proposed Approach}
    \subsection{Problem Formulation}
    \label{subsection:problem_formulation}
    Given a number of $N$ consecutive timestamps $Ts_1, Ts_2, ..., Ts_N$, representing the day hours in our use case, the aim is to determine a strategy that assigns the grid topology to adopt at each timestamp. In this context, we use the terms "strategy" and "plan" interchangeably. The goal is twofold: (i) to optimise the maximum load flow under the N-1 contingency condition throughout the entire plan, denoted as $\max\rho_{n-1}$, and (ii) the frequency of topology changes within a plan (i.e., the number of times the current grid topology is changed), referred to as "N switching". These correspond to objectives (2) and (3) outlined in Section~\ref{section:background} respectively. With MOO, we seek to generate plans that approximate the PF, capturing various trade-offs between these two objectives.

    \subsection{Multi-Objective Optimisation Approach}
        Our approach overcomes the inefficiencies of state-of-the-art MORL algorithms, which arise from the extensive search space required to explore multiple trade-offs during training in conjunction with large state and action spaces. To address this issue, we adopt a two-phase methodology that separates RL training and planning.
    
        \paragraph{\textbf{Phase 1: Training an RL Agent}} In the first phase, we train an RL agent to optimise a \textit{utility reward function}\footnote{A utility reward function in MOO is a scalarisation function that combines multiple objectives into a single numerical value.} that encapsulates the objectives of interest, specified to reflect the importance of each objective and to optimise a pre-defined trade-off among them.
        
        \paragraph{\textbf{Phase 2: Plan Generation using Single-Step Planning (SSP)}} In the second phase, the trained RL agent is deployed to generate a set of \textit{non-dominated set}\footnote{A non-dominated set in multi-objective optimisation comprises solutions where no single solution is superior in all objectives compared to another.} of plans that form an approximation of the PF. First, the agent is deployed at each timestamp by observing the state of the grid and generating the grid topology to adopt based on the learned policy. The generated topology is supposed to optimise both objectives defined in subsection~\ref{subsection:problem_formulation}. Second, for each value of "N switching" objective, which takes discrete values, the plan generation algorithm seeks to find the plan that achieves the best possible value of N-1 load objective. For example, when "N switching" = 1, the algorithm searches for the specific timestamp to change the grid topology (as recommended by the agent) to optimise the N-1 load objective over the entire plan. This process is repeated for all other values of "N switching", which typically range from 1 to 24. The algorithm filters the generated plans by eliminating dominated solutions, leaving a refined set that represents the PF approximation. A structured description of the algorithm is given in Alg.~\ref{alg:plan_generation_algo} in the Appendix.

    \subsection{Our methods}
        For phase 1, we propose two distinct RL agents, while phase 2 remains consistent across all experiments.
        
        \paragraph{\textbf{Single-Step Agent (SSA)}} SSA is trained to act on each timestamp independently. For a timestamp $Ts_i$, the agent decides which topology $Topo_i$ to put in place by sequentially acting on different substations. The agent is rewarded only at the end of the episode, once the final topology is formed. The reward, is provided only in the final state, and is defined as: $ R\left(S_i^F\right) = w_1 \times R_1 + w_2 \times R_2 + w_3 \times R_3$, where:
        
        \begin{itemize}
            \item $S_i^F$ is the final state after simulating the obtained topology $Topo_i$.
            \item $ R_1 = f_1(\max\rho_{n-1}^i) $, minimises the maximum N-1 load at the current timestamp $Ts_i$. $f_1$ can be any monotonically decreasing function. 
            \item $ R_2 = | \Omega_s |$, where $\Omega_s$ is the set of the next consecutive timestamps where the topology $Topo_i$ is stable. A topology $Topo_i$ is stable with respect to a given timestamp $Ts_j$ if when simulated with its injections $I_j$, $\max\rho_{n-1}^j < 1$.
            \item $ R_3 = f_3(\text{Aggr}_{j \in \Omega_s} \rho_{n-1}^j ) $, minimises the aggregated values of $\max\rho_{n-1}$ across the set of stable timestamps in $\Omega_s$. We use the $max$ aggregation function in our experiments. $f3$ can be any monotonically decreasing function.
        \end{itemize}

        The weights $w_1$, $w_2$, and $w_3$ determine the relative importance of $R_1$, $R_2$, and $R_3$, respectively, in the overall reward function. $R_1$ optimises the N-1 load objective. "N switching" objective is optimised by $R_2$ and $R_3$. The weights were set to $(w_1, w_2, w_3) = (0.15, 0.7, 0.15)$ in our experiments, and the RL agent was trained using the Proximal Policy Optimization (PPO) algorithm \cite{schulmanProximalPolicyOptimization2017}.
        
        \paragraph{\textbf{AlphaZero Agent (AZA)}} AlphaZero Agent \cite{silverMasteringChessShogi2017, dorferPowerGridCongestion2022} acts on the sequence of timestamps of the whole day. It is trained to provide plans that minimise $\max\rho_{n-1}$ and the amount of switching. In contrast to SSA, which is trained on timestamps independently, AZA produces a plan deciding which topology to apply on each timestamp of the day. For each timestamp, two rewards are computed, $R_1 = f_1(\max\rho_{n-1})$, and $R_2 = 1$ if the current topology is the same as in the previous timestamp or $R_2 = 0$ otherwise. $f_1$ is a monotonically decreasing function. The agent is trained on a \textit{utility reward function} defined as $R = w_1 \times R_1 + w_2 \times R_2$. $(w_1, w_2) = (0.95, 0.05)$ were used in our experiments. Once trained, we deploy the obtained policy on each timestamp following phase 2 algorithm of our approach to generate \textit{non-dominated set} of plans. 

        \smallskip
        Combined with phase 2 planning, the two methods described above are referred to as \textit{SSA + SSP} and \textit{AZA + SSP} respectively.

    \subsection{Baselines}
        The following baselines are used for comparison:
    
        \textbf{Reference Strategy:} Place all transmission lines on \textit{bus A} at every substation for all timestamps.
        
        \textbf{Expert Strategy 1:} Adopt a fixed operator topology that changes only substation $0$ to reroute power flows towards a targeted substation. This topology is maintained throughout the day.
        
        \textbf{Expert Strategy 2:} Adopt a fixed operator topology that disables a specific line to change the distribution of power flows. This is applied to all timestamps of the day.
        
        \textbf{Expert Strategy 1\&2:} Combine strategies 1 and 2 described above.
        
        \textbf{Expert Set:} Includes the reference and all expert strategies described above.

\begin{figure}[t]
  \centering
  \includegraphics[width=0.7\linewidth]{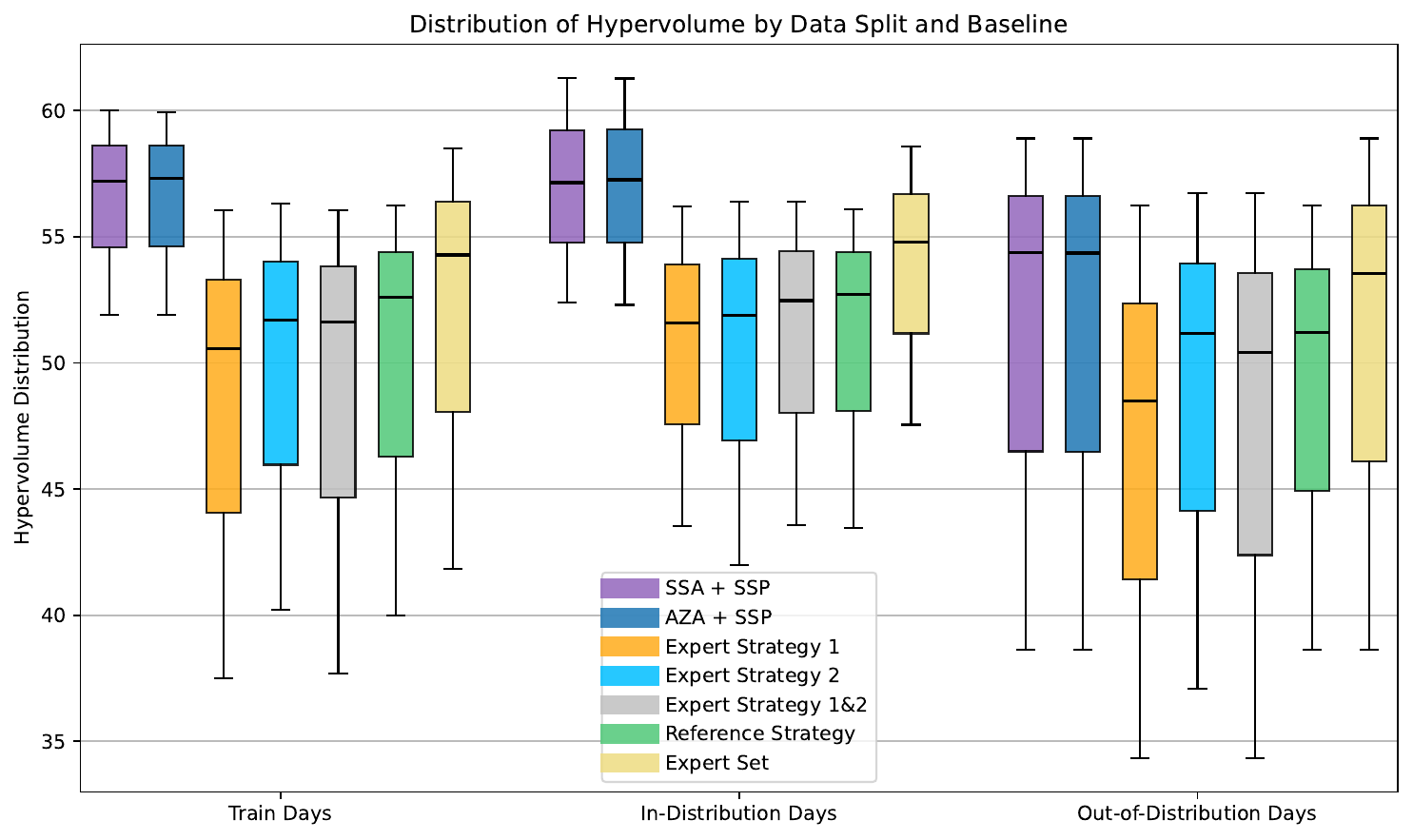}
  \caption{Distribution of hypervolume values by approach and baseline across the three data splits. Higher values indicate better performance.}
  \label{fig:hypervolume}
\end{figure}

\begin{figure*}[ht] 
    \centering
    \begin{subfigure}[t]{0.48\linewidth} 
        \includegraphics[width=\textwidth]{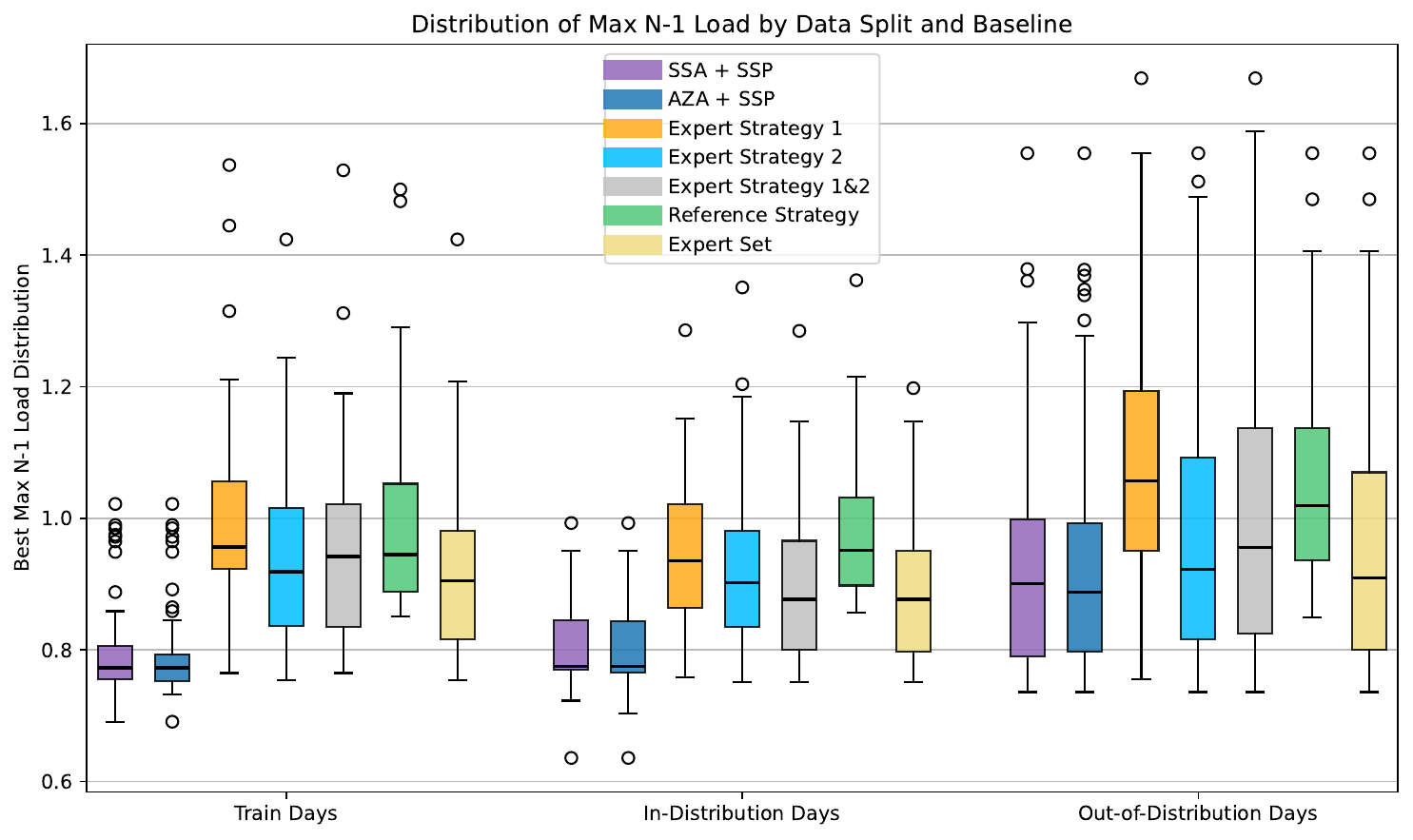} 
        \caption{Distribution of $\max\rho_{n-1}$ by approach and baseline across the three data splits.}
        \label{subfig:best_max_rho_n-1}
    \end{subfigure}
    \begin{subfigure}[t]{0.48\linewidth}
        \includegraphics[width=\textwidth]{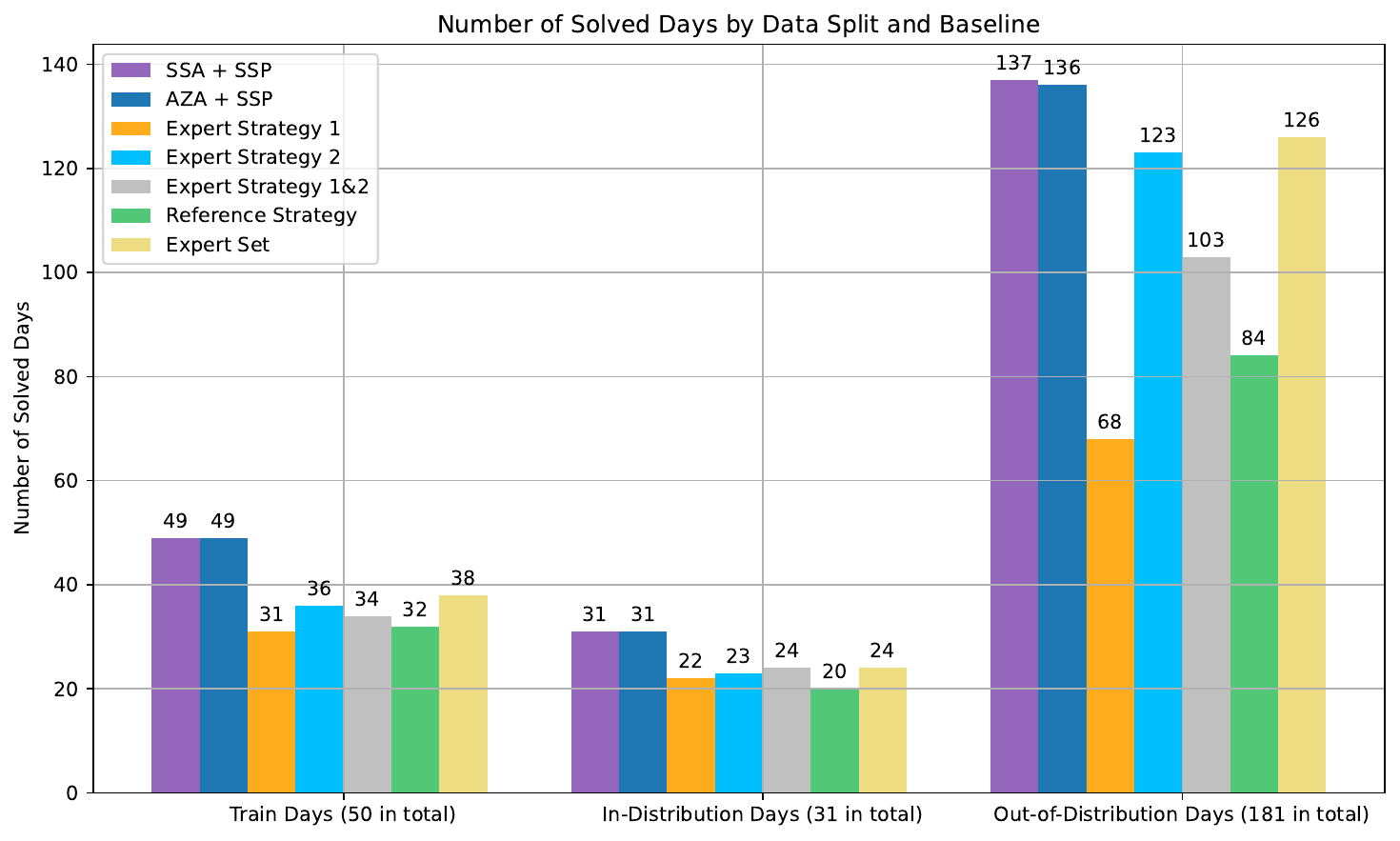}
        \caption{Number of solved days by approach and baseline for each data split. A day is considered solved if at least one of the produced plans brings $\max\rho_{n-1}$ under the threshold of 1.}
        \label{subfig:n_solved_days}
    \end{subfigure}
    \caption{Statistics considering the best plan that yields the best $
    \max\rho_{n-1}$. }
    \label{fig:best_max_n_1_and_n_solved_days}
\end{figure*}

\section{Experimental Setup}

    \paragraph{\textbf{Real-World Data}} The experimental setup is based on the TenneT use case described in \cite{viebahnGridOptionsToolRealWorld2024} and real-world data providing historic detailed information about the power grid for a set of days. This data includes sequences of timestamps, where each timestamp corresponds to one hour of the day. For every hour, the dataset specifies the load demand and the generator production values at all substations. In addition, the base grid topology, which outlines the connectivity of substations and transmission lines, is provided along with the physical properties of the lines, such as their capacity and thermal limits, necessary for simulation. The grid comprises 1659 substations, 1338 power lines and 1716 injections.

    \paragraph{\textbf{Environment Setup}} The simulation environment is designed to approximate real-world grid operations. It uses a DC flow approximation of the AC non-linear power flow equations to model the grid dynamics \cite{stottReviewLoadflowCalculation1974, vandijkBusSplitDistribution2024}. This simplification provides computational efficiency while maintaining sufficient accuracy for evaluating grid performance under various scenarios.
    
    \textit{The state space} is constructed from a combination of grid information, including the current state of the grid topology and the line flows. This information enables the RL agent to make informed decisions at each step of the decision-making process.
    
    \textit{The action space} is derived from a set of predefined target topologies provided in the real-world data. These target topologies represent potential grid configurations that meet operational constraints. The total number of target topologies is approximately $1e5$. Each target topology is decomposed into a sequence of unitary actions. Each unitary action modifies the configuration of a single substation at a time, allowing the agent to incrementally transition from the reference grid state (i.e. all branches are connected to \textit{bus A}) to the target topology. This decomposition technique originally proposed in \cite{dorferPowerGridCongestion2022}, allows the RL agent to operate efficiently within the large and complex action space. By design of the set of target topologies, the topological depth is restricted to a maximum of three. This constraint ensures that no more than three substations can be manipulated in a single sequence. The target topologies specify only the branch topology, indicating which bus within a substation a line connects to. The injection topology, indicating where injections are connected, is obtained by manual optimisation to determine the configuration that maximises the agent's reward.
    
    \paragraph{\textbf{Evaluation}} The dataset is split into two consecutive segments: the first spans a continuous time period, while the second follows immediately after. From the first segment, we randomly selected 50 days for training and 31 days for in-distribution evaluation. The second segment, which occurs later in time, is allocated for out-of-distribution evaluation, consisting of 181 days. It is labelled as out-of-distribution because data from future periods is expected to deviate considerably from the training data. To assess our methods and baselines, we calculate the hypervolume of the plans provided for each day and analyse its distribution over the data splits. The hypervolume is computed relative to a fixed reference point: ($\max\rho_{n-1}$, "N switching") = (3.1, 25) in our evaluation. We also explore the distribution of the best $\max\rho_{n-1}$ values achieved for each day.

\section{Results and Discussion}

    To contextualise the performance of our methods and the expert baselines, it is important to note that \textit{SSA + SSP} and \textit{AZA + SSP} are trained to adapt to different power grid scenarios, producing various topologies and plans. In contrast, the expert baselines rely on fixed topologies that often work well but cannot adjust to new or varying conditions. 

    Fig.~\ref{fig:hypervolume} presents the hypervolume variability and median for each baseline across the three data splits. \textit{SSA + SSP} and \textit{AZA + SSP} achieve the highest median hypervolumes overall, while also maintaining superior minimum values compared to the expert baselines. Our methods demonstrate comparable hypervolume ranges across the training and in-distribution splits, suggesting stable performance on data closely resembling the training data. However, the results on out-of-distribution split exhibits noticeable degradation, which can be attributed to the agents' limited exposure to such scenarios during training. 

    As shown in Fig.~\ref{subfig:best_max_rho_n-1}, our approaches consistently yield lower $\max\rho_{n-1}$ values across the various data splits. While the expert baselines perform well much of the time, they lack flexibility to handle different scenarios. In contrast, \textit{SSA + SSP} and \textit{AZA + SSP} adapt to the different scenarios based on their training phase, achieving better overall distributions of $\max\rho_{n-1}$. 

    From the perspective of TSOs, a “solved day” is one where the N-1 load flow remains below 1.0. As illustrated in Fig.~\ref{subfig:n_solved_days}, our methods solve all training days with the exception of one outlier, and achieve 100\% success rate for in-distribution days. In out-of-distribution scenarios, they solve more than 75\% of days, surpassing the expert set baseline’s ability to maintain reliable N-1 load flows. 
 
    Overall, \textit{SSA + SSP} and \textit{AZA + SSP} demonstrate robust adaptability and superior results across all three metrics—hypervolume, $\max\rho_{n-1}$, and number of solved days—when compared to expert baselines. Their ability to learn from data and produce multiple solutions per day leads to consistently better outcomes in both familiar and novel scenarios.

\section{Conclusion and Future Work}

    In this work, we demonstrated promising results for day-ahead planning on real-world data from a use-case defined by TenneT TSO to control a large grid network with hundreds of substations and power lines. Our findings confirm the potential of RL for optimising power grid operations, as highlighted in prior literature. Building on this foundation, we introduced a multi-objective approach that generates an efficient approximation of the PF for optimising multiple objectives. 
    
    The proposed approach involves a two-phase process: first, training an RL agent to learn an optimal policy for a predefined utility function with the ability to handle the complex grids; then, employing a multi-objective optimisation planning algorithm to approximate the PF. This approach yields strong qualitative results, outperforming expert solutions, with inference time reduced to just a few minutes, making it well-suited for practical, real-world day-ahead planning, where computational efficiency is critical.  
    
    While this work focused on optimising two objectives, the framework can be extended to include additional objectives. For instance, the topological depth objective (objective (4) in Section~\ref{section:background}) could be incorporated by conditioning the RL policy on the discrete specific values of topological depth. The planning algorithm could then generate distinct solutions conditioned on these values, producing an approximation of the PF for each specified topological depth. By merging these individual Pareto Fronts and applying PF filtering, a global PF could be derived. 
    
    Another compelling direction for future work is to generate diverse plans that represent the same point on the PF. This would offer operators a wider range of options, empowering them to choose solutions tailored to particular operational needs or situational constraints. 

\bibliographystyle{unsrt}  
\bibliography{references}  

\newpage
\appendix
    \section{Appendix}
    
    \begin{figure*}[h]
      \centering
      \includegraphics[width=0.98\linewidth]{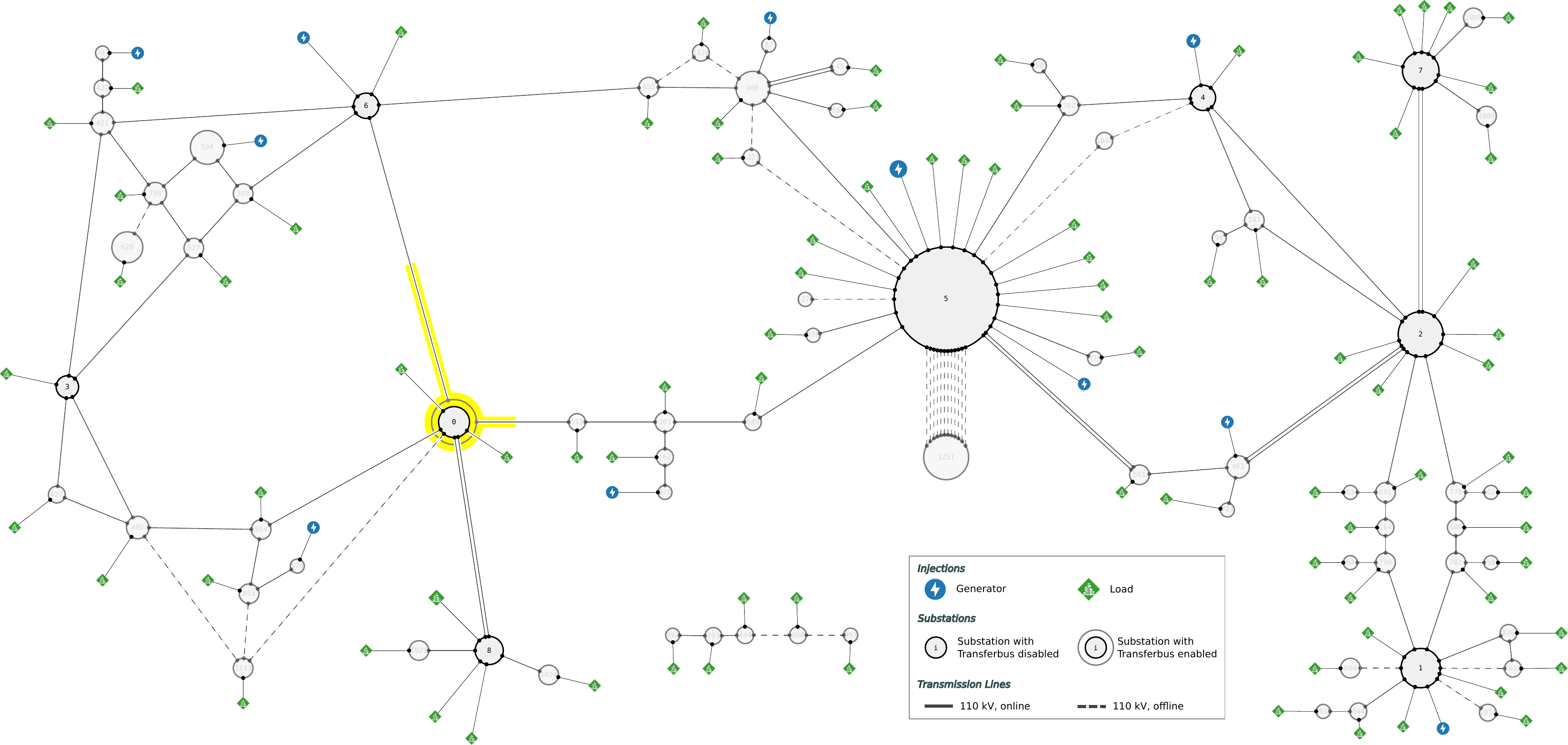}
      \caption{Depiction of part of the grid used in our experiments. The entire grid contains 1659 substations, 1338 power lines and 1716 injections.}
      \label{fig:grid_network}
    \end{figure*}
    
    \subsection{Number of switching timestamps results}
\begin{figure}[h]
          \centering
          \includegraphics[width=0.7\linewidth]{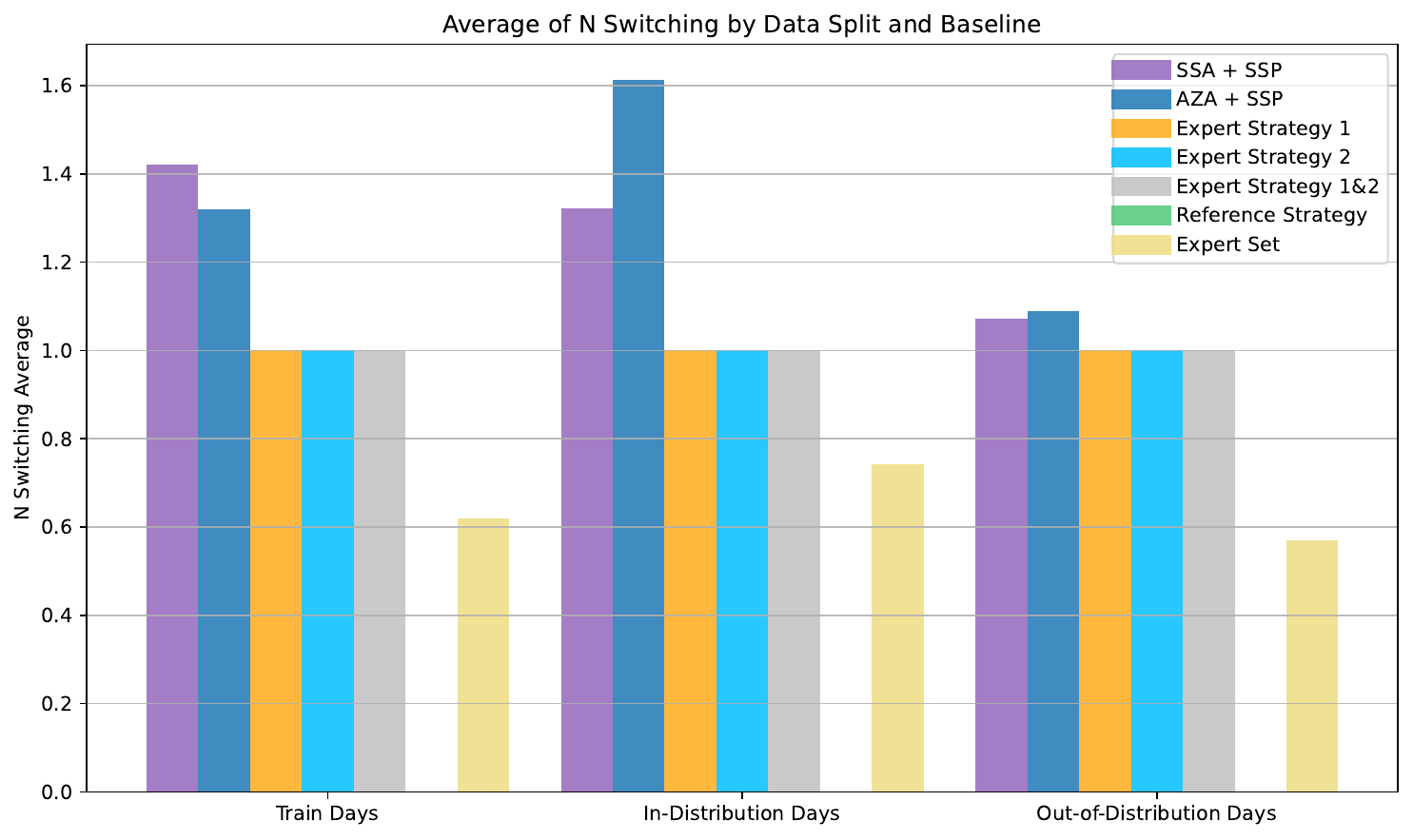}
          \caption{Corresponding "N switching" objective average for best plans with best $\max\rho_{n-1}$ depicted in Fig.\ref{fig:best_max_n_1_and_n_solved_days}.}
          \label{fig:corresp_average_n_switching_ts}
        \end{figure}

        Fig.~\ref{fig:corresp_average_n_switching_ts} shows the "N switching" objective average for the best plans achieving the highest $\max\rho_{n-1}$ presented in the results section. The expert set baseline exhibits the lowest "N switching" average, with a value below one, indicating that the reference strategy often provides the optimal $\max\rho_{n-1}$, which other expert topologies generally fail to surpass. Our methods involve slightly more switching but remain moderate with an average less than two.

    \subsection{Acronyms}
    The acronyms used in the paper are summarised in Table~\ref{table:acronyms}.

    \begin{table}[h]
        \centering
        \begin{tabular}{|l|l|}
        \hline
        \textbf{Acronym} & \textbf{Full Form} \\ \hline
        TSO                  & Transmission System Operator     \\ \hline
        RL                   & Reinforcement Learning            \\ \hline
        MORL                 & Multi-Objective Reinforcement Learning \\ \hline
        MOO                  & Multi-Objective Optimisation      \\ \hline
        PF                   & Pareto Front                      \\ \hline
        PPO                  & Proximal Policy Optimisation      \\ \hline
        AZA                  & AlphaZero Agent                   \\ \hline
        SSA                  & Single-Step Agent                 \\ \hline
        SSP                  & Single-Step Planning              \\ \hline
        
        \end{tabular}
        \caption{Acronyms and their Full Forms}
        \label{table:acronyms}
    \end{table}

    \subsection{Grid Architecture}
        The power grid used in our use case, a portion of which is depicted in Fig.~\ref{fig:grid_network}, operates as a complex network of interconnected elements, each playing a role in ensuring the generation, transmission, and distribution of electricity. Key elements of the grid include:

        \begin{itemize}
            \item \textbf{Injections (Loads and Generators):} Injections are points where power is either supplied to or drawn from the grid. \textbf{Generators} inject power into the grid by converting various energy sources (e.g. fossil fuels, wind, solar) into electricity, while \textbf{loads} consume power to meet the demands of industries and households.
            \item \textbf{Substations:} Substations serve as hubs in the grid where power is transformed, or rerouted. Each substation is equipped with two buses ($A$ and $B$) that serve as connection points where transmission lines, generators and loads are linked. They allow reconfiguring the grid topology by switching by switching lines between buses to adapt to changing conditions. Bus $B$ is also called \textit{transferbus}.
            \item \textbf{Transmission lines:} Transmission lines are the pathways that carry electricity between substations. They can be \textit{online}, actively carrying power, or \textit{offline}, temporarily out of service due to maintenance.
        \end{itemize}

    \subsection{Planning Algorithm}
    The planning algorithm is outlined in Alg.~\ref{alg:plan_generation_algo}.

    \begin{algorithm}[b!]
        \SetAlgoLined
        \KwIn{
            \begin{itemize}
                \item Trained RL agent with policy $\pi$ deployable at any timestamp.
                \item Discrete "N switching" values $N_s \in \{1, \ldots, 24\}$.
            \end{itemize}
        }
        \KwOut{Set of non-dominated plans approximating the Pareto Front.}
        \BlankLine  
        
        \For{each timestamp $Ts$ of the day}{
            Observe the grid state $s_0$ in the reference topology\;
            Construct the topology suggested by the agent by sequentially querying the policy $\pi$\;
            Simulate the obtained topology with subsequent timestamps \;
        }
        
        \BlankLine 
        Initialise the solution set $S$ with the reference plan\;
        \For{$N_s \in \{1, \dots, 24\}$}{
            Generate all combinations of timestamps $\{t_1, \dots, t_{N_s}\}$ for topology changes and construct plans\;
            Select plan $P_{N_s}$ with the best $\max\rho_{n-1}$ \;
            Add $P_N$ to $S$\;
        }
        Remove dominated plans from $S$\;
        \Return $S$\;
        \caption{Phase 2, Single-Step Planning Algorithm for Optimising $\max\rho_{n-1}$ and "N switching" objectives} 
        \label{alg:plan_generation_algo}
    \end{algorithm}

\end{document}